# Deep Reinforcement Learning for Artificial Upwelling Energy Management

Yiyuan Zhang [1], Wei Fan [1]

**Abstract**: The potential of artificial upwelling (AU) as a means of lifting nutrient-rich bottom water to the surface, stimulating seaweed growth, and consequently enhancing ocean carbon sequestration, has been gaining increasing attention in recent years. This has led to the development of the first solar-powered and air-lifted AU system (AUS) in China. However, efficient scheduling of air injection systems in complex marine environments remains a crucial challenge in operating AUS, as it holds the potential to significantly improve energy efficiency. To tackle this challenge, we propose a novel energy management approach that utilizes deep reinforcement learning (DRL) algorithm to develop efficient strategies for operating AUS. Specifically, we formulate the problem of maximizing energy efficiency of AUS as a Markov decision process and integrate the quantile network in distributional reinforcement learning (QR-DQN) with the deep dueling network to solve it. Through extensive simulations, we evaluate the performance of our algorithm and demonstrate its superior effectiveness over traditional rule-based approaches and other DRL algorithms in reducing energy wastage while ensuring the stable and efficient operation of AUS. Our findings suggest that a DRL-based approach offers a promising way to improve the energy efficiency of AUS and enhance the sustainability of seaweed cultivation and carbon sequestration in the ocean.

[1] Ocean College, Zhejiang University, Zhoushan 316000, China

# 1 Introduction

Artificial upwelling (AU), as a form of geoengineering, has garnered significant attention due to its potential to enhance ocean primary productivity and mitigate atmospheric $CO_2$ accumulation in a sustainable manner [1,2]. The process of AU involves transporting nutrient-rich bottom water to the surface, thereby creating an environment conducive to the growth of phytoplankton and other marine organisms [3,4]. This, in turn, can lead to the formation of larger and more efficient ocean carbon sinks [2,5]. As a result, AU has been considered a promising tool to help the Earth heal itself [3,6]. Among the various upwelling techniques, air-lifted AU has emerged as a particularly efficient method [7,8], which generates a bubble-entrained plume (BEP) by injecting air into an underwater nozzle [8,9], as illustrated in Figure 1.

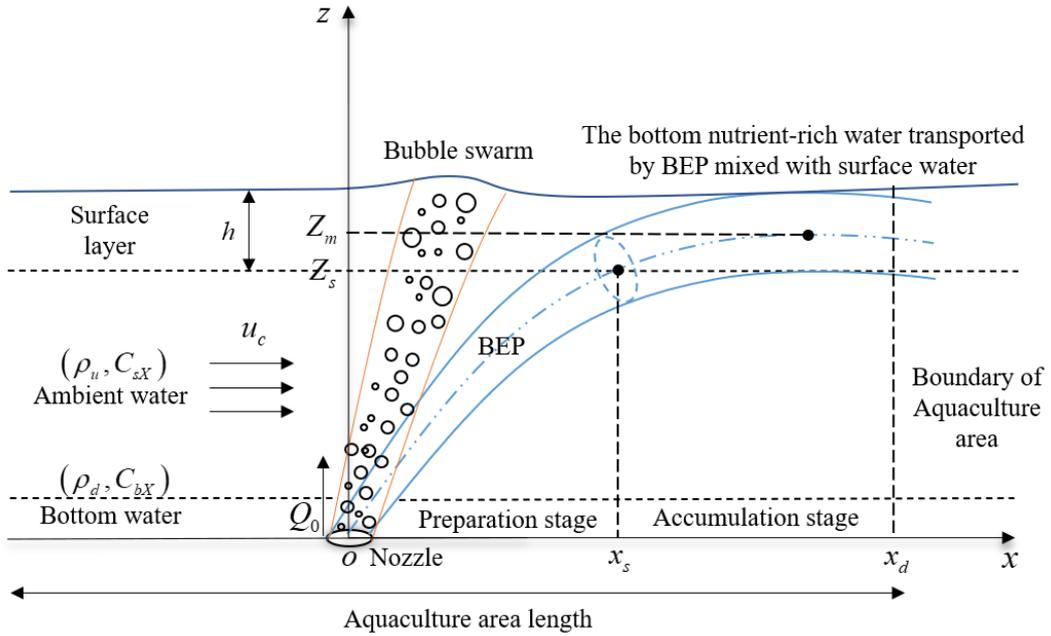

Figure 1. Transportation of nutrient-rich bottom water through air-lifted artificial upwelling. Under crossflow, the plume initially separates from the bubble swarm, then enters the surface layer (depth < 2m), and subsequently undergoes mixing with the surface water. (Surface water refers to the ambient water present in the surface layer.)

However, the energy-intensive process of lifting nutrient-rich bottom water to the surface incurs high energy expenditures. This presents a significant challenge for the implementation of large-scale AU field applications in the face of global energy scarcity, given the limited availability of energy resources in open seas and the exorbitant costs associated with land-based electricity [7]. Address this issue, Fan. et al [10] constructed the first solar-powered and air-lifted artificial upwelling system (AUS) in Aoshan Bay, China, as shown in Figure 2. This innovative system is able to significantly increase the biomass and carbon removal of cultivated seaweed [10], demonstrating the feasibility of using renewable energy to power large-scale AU systems.

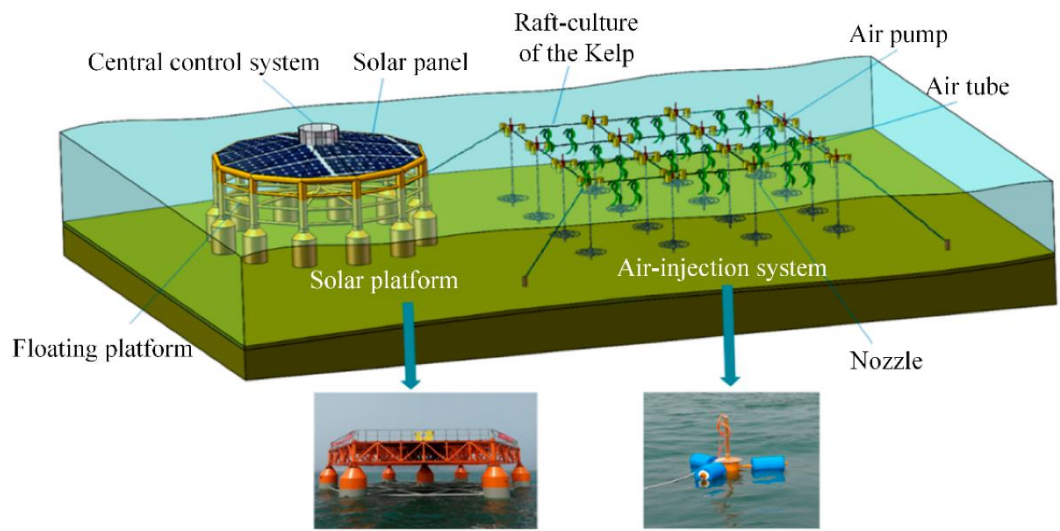

Figure 2. Schematic diagram of the air-lifted artificial upwelling system in Aoshan bay [11].

Considering the variability and unknown nature of renewable energy [12], it is necessary to develop a dependable and efficient artificial upwelling energy management system (AUEMS). This system should ensure the stable long-term operation of essential loads (i.e., sensors and CPUs), while also optimizing the timing of air injection within the AUS to maximize nutrient transport from bottom water to the seaweed during periods of peak photosynthetic activity, thereby enhancing productivity. However, achieving this objective poses significant challenges. Firstly, the performance of the AUS is influenced by various factors, including the dynamic nature of the surrounding environment (e.g., tides, temperature, and light) and the operational state of the system. Secondly, acquiring the statistical distribution of all possible combinations of stochastic system parameters proves to be a formidable task.

Additionally, there exists a temporal coupling constraint between the energy storage system (ESS) and the air injection system, whereby current actions have repercussions on future decisions. To this end, Lin et al. [13] proposed a rule-based method to optimize injection patterns of AUS based on environmental factors in Aoshan Bay, but failed to consider the impact of the ESS. Moreover, conventional rule-based and model-based strategies tend to be inefficient in practice, owing to the intricate nature of ocean hydrodynamics and environmental disturbances [14].

In recent years, deep reinforcement learning (DRL) techniques have emerged as powerful methods for solving complex control problems, achieving remarkable success in playing Atari and Go games [15,16]. By combining deep learning with reinforcement learning (RL), DRL techniques can overcome the limitations of conventional reinforcement learning approaches by building deep neural networks to estimate values, correlate state-action pairs, and handle large state spaces [17]. Notably, DRL-based energy management approaches have been increasingly utilized in various fields, including energy management of hybrid energy systems and building air conditioning systems [12,18]. However, to the best of the authors' knowledge, few research has yet applied DRL techniques to energy management for AU and this work represents the first attempt.

In this work, we focus on the energy optimization problem for an AUS with photovoltaic generation system, ESS, air injection system, and essential loads. Our primary aim is to maximize the AUS energy efficiency over a defined time frame (e.g., a seaweed culture cycle) while securing the continuous power supply of essential loads. To achieve this, we propose a distributional reinforcement learning with quantile regression (QR-DQN) based energy management algorithm. This algorithm enables us to determine the input power of the air injection system solely based on the current observation information, simplifying the decision-making process.

The organization of this paper is as follows: Section II delineates the system model and problem formulation. In Section III, we present a detailed description of the energy management algorithm for the AUS. The effectiveness of the algorithm is then evaluated through simulation results in Section IV. Finally, we present our conclusions and discuss future work in Section V.

## 2 Problem Modelling

As illustrated in Figure 3, our study focuses on an AUS consisting of photovoltaic generation systems, ESS, loads, and AUEMS. The ESS is equipped with lithium batteries that store energy locally and power loads when needed. The loads in the AUS include the essential loads (e.g., sensors and CPUs) and working loads (e.g., air compressors). For the AUEMS, it is necessary to ensure that the long-term power requirements of the essential loads are met. The tasks of the air injection can be scheduled at appropriate times and its switching and input power can be regulated by adhering to specific operational requirements, thereby enabling flexible adjustment of its operation time and energy consumption.

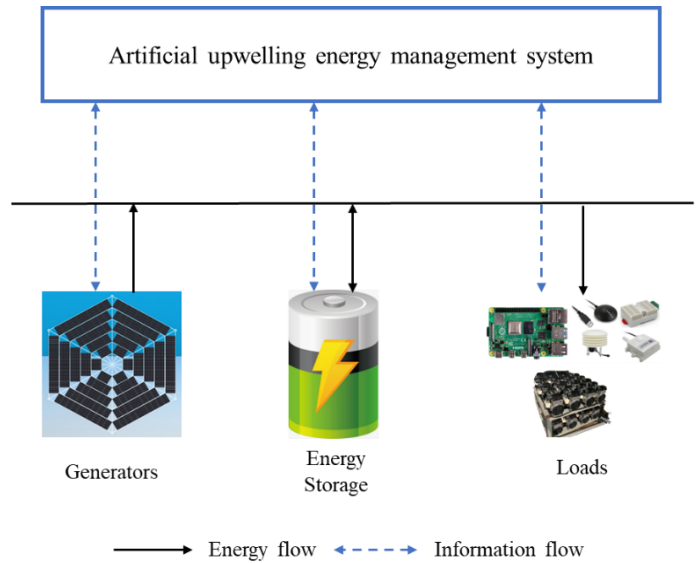

Figure 3. Energy exchange in the artificial upwelling system

Assumed that the AUS operates in discrete time periods, i.e., $t \in [0, K]$, where $K$ represents the total number of time periods. For simplicity, we normalize the duration of each time period $\Delta t$ to a unit time (e.g., one hour) for power and energy equivalence. Within each time period, the AUEMS continuously decides on the ON-OFF and input power of the air injection system based on a set of observations (e.g., temperature, light intensity, and the energy level of ESS). In this section, we developed the models related to ESS and AUEMS, and formulated the problem of maximizing energy efficiency for AUS. Since directly solving the efficiency maximization problem is challenging, we redefined it as a Markov Decision Process (MDP).

## 2.1 System Model

(1) Energy Storage Model

Let $E_t$ denote the stored energy of ESS at time period $t$, and the kinetic model of ESS energy storage is given by [18]:

$$E_{t+1} = E_t + \eta_c c_t - \frac{d_t}{\eta_d} \tag{1}$$

where $\eta_c \in (0, 1]$ and $\eta_d \in (0, 1]$ represent the charge and discharge efficiency coefficients; $c_t$ and $d_t$ denote the charging power and discharging power of ESS, respectively. Additionally, the energy storage $E_t$ and charging and discharging power of ESS $c_t / d_t$ are subject to the following constraints:

$$\begin{cases} E_{\min} \leq E_t \leq E_{\max} \\ 0 \leq c_t \leq c_{\max} \\ 0 \leq d_t \leq d_{\max} \end{cases} \tag{2}$$

where, $E_{\max}$ and $E_{\min}$ denote the upper and lower limits of the stored energy of the ESS, respectively, while $c_{\max}$ and $d_{\max}$ denote the maximum charge and discharge power of the ESS, respectively. To prevent concurrent charging and discharging of the ESS:

$$c_t \cdot d_t = 0 \tag{3}$$

(2) Air Injection Model

The air injection system consists of an arrangement of air pipes, nozzles and air compressors. Its function is to inject air into the underwater nozzles through pipeline laid on the seabed, thereby generating AU. The power consumption of the air injection system at time period $t$ is determined by the number of operating air compressors $m_t$ and the rated power $p_0$. Furthermore, the total output air injection rate $Q_t$ is determined by $m_t$ and the rated air injection rate of $Q_0$.

$$q_t = m_t \times p_0, m_t \in [0, M] \tag{4}$$

$$Q_t = m_t \times Q_0, m_t \in [0, M] \tag{5}$$

where $M$ represents the total number of air compressors.

(3) Power Generation Model

Assuming that all the energy for AUS is derived from solar generation, the amount of power generated at time period $t$ can be expressed as [13]:

$$g_t = \frac{H_t \times P_{AZ} \times K}{1000 \text{W}/\text{m}^2} \tag{6}$$

where $H_t$ represents the total solar irradiance during this period; $P_{AZ}$ is the installed capacity of the system and $K$ is the correction factor, which accounts for factors such as line consumption, surface contamination, and the angle of the solar power panels.

$$H_t = \int I_t dt = I_t \cdot \Delta t \tag{7}$$

where $I_t$ denotes the light intensity at time period $t$, W/m$^2$.

(4) Generation Balance

To maintain the power balance in the power grid of AUS, the total power supply must be equal to the power demand of all loads:

$$g_t + d_t = b_t + q_t + c_t \tag{8}$$

where $b_t$ is the power demand of essential loads.

(5) Artificial Upwelling Transport Model

To evaluate the performance of the AUS at time period $t$, a key metric to consider, namely the cumulative volume of nutrient-rich bottom water transported by each nozzle to the surface water within a given area (e.g., a seaweed culture farm, see Figure 2) [19]. This metric is expressing in the Eq. (9). It is important to note that successful upwelling relies on certain conditions being met. Specifically, if the maximum height of the BEP, denoted as $Z_m$, is lower than the height of the surface layer ($Z_s$), or if the lateral displacement of the BEP as it reaches the surface layer ($x_s$) exceeds the lateral distance between the nozzle and the boundary of the aquaculture area ($x_d$), then the bottom water cannot be effectively transported to the surface [11].

$$V_t = \begin{cases} 0, & \text{if } Z_{m,t} < Z_{s,t} \text{ or } x_{s,t} > x_d \\ \sum_{i=1}^{N} V_{i,t} = \sum_{i=1}^{N} f_{1,i,t} Q_{w,i,t}, & \text{otherwise} \end{cases} \quad (9)$$

where $N$ is the number of nozzles, $f_{1,i,t}$ represents crossflow effect factor, and $Q_{w,i,t}$ denotes the volume flow rate of transporting bottom water by the $i$-th nozzle at time period $t$, which can be estimated by approximating the volume flow rate of BEP at the nozzle outlet [9]:

$$Q_{w,i,t} = 0.06 Q_{i,t}^{1/3} \Delta z^{5/3} \left( \tanh\left( \frac{\sqrt[3]{gQ_{i,t}}}{H_0 \times 0.25} \right) \right)^{3/8} \quad (10)$$

where $Q_{i,t}$ represents the volume flow rate of injected air, and $\Delta z$ symbolizes the virtual displacement of the nozzle after considering the effect of pressure drop effect [20]:

$$\Delta z = \frac{d_0}{1.2 \times \alpha} \quad (11)$$

where $d_0$ refers to the diameter of the nozzle, while $\alpha$ represents the entrainment coefficient, which can be calculated according to Kobus et al. [21]:

$$\alpha = 0.082 \left[ \tanh\left( \frac{\sqrt[3]{gQ_{i,t}/H_0}}{v_s} \right) \right]^{3/8} \quad (12)$$

where $H_0$ corresponds to the head height at standard atmospheric pressure (generally 10.4 m); $v_s$ denotes the bubble slip velocity (estimated to be 0.3 m/s [9]), and $g$ represents the acceleration of gravity.

The maximum height of the BEP $Z_{m,t}$ is dependent on the crossflow velocity $u_{c,t}$, the air injection rate $Q_{0,t}$, and the density difference between the upper and bottom water layers $\Delta\rho$, which can be expressed by the empirical formula [11,19]:

$$Z_{m,t} = 0.0006 \times \frac{(\alpha^{-2/3} \cdot Q_{0,t}^{1/3})^{1.9}}{\Delta\rho^{0.4} \cdot u_{c,t}^{1.1}} - 0.4446 \quad (13)$$

Crossflow effect factor $f_{1,i,t}$ represents the impact of ocean currents and nozzle location on the performance of AUS [19]

$$f_{1,i,t} = \left(\frac{x_{d,i,t} - x_{s,i,t}}{|u_{c,t}| \cdot \Delta t}\right)^2 \left(\frac{|u_{c,t}| \cdot \Delta t - x_{d,i,t}}{x_d - x_{s,i,t}} + \left|\frac{u_{c,t-1}}{u_{c,t}}\right|\left(\frac{dir \cdot (x_{aq} - x_{d,i})}{(x_{d,i,t} - x_{s,i,t-1})} + \frac{1}{2}\right) + \frac{1}{2}\right) \quad (14)$$

where $x_{aq}$ is the length of the aquaculture farm and $dir \in \{0, 1\}$ denotes the direction factor. When $dir = 1$, it signifies that there has been a change in the crossflow direction in the subsequent time period $t+1$.

(6) Total Efficiency Maximum Problem

Given that the energy generation is constant for defined time frame, maximizing the total energy efficiency during those periods is equivalent to maximizing the cumulative volume of nutrient-rich bottom water transport. Based on the above model, the total efficiency maximum problem can be described as follows:

$$\textbf{(P1)} \quad \max \sum_{t=1}^{T} \mathbb{E}\{f_{2,t} V_t\} = \max \sum_{t=1}^{T} \mathbb{E}\left\{f_{2,t} \sum_{i=1}^{N} f_{1,i,t} Q_{w,i,t}\right\} \quad (15)$$
$$s.t. \ (1)-(8)$$

where the expectation operator $\mathbb{E}$ signifies the stochastic nature of the system parameters (i.e., renewable energy generation output $g_t$, temperature, and tidal variations) and the possible stochastic control actions at each time period (the number of air compressors turned on). Since the main function of AUS is to enhance photosynthesis in seaweed and promote carbon sequestration, the photosynthetic factor $f_{2,t}$ is introduced to account for the effect of temperature and light on seaweed photosynthetic rate [22]:

$$f_{2,t} = f(I_t) \times f(T) = \frac{I_t}{I_s} \times \exp\left(1 - \frac{I_t}{I_s} - 2.3 \times \left(\frac{T_t - T_{opt}}{T_x - T_{opt}}\right)^2\right) \quad (16)$$

where $I_s$ represents the optimal light intensity for the growth of seaweed growth. $T_{opt}$ represents the optimum growth temperature; $T_x$ represents the temperature ecological amplitude. If $T \leq T_{opt}$, $T_x = T_{min}$, otherwise $T_x = T_{max}$. $T_{min}$ and $T_{max}$ are the lower and upper temperature limits.

The challenges associated with solving Problem **P1** have been discussed in Section 1 of this study. In order to address these challenges, we propose a novel method that allows for the solution of **P1** without relying on prior knowledge of the environmental dynamics or the stochastic system parameters. This is accomplished by reformulating the problem of maximizing efficiency as a Markov Decision Process (MDP) problem.

## 2.2 MDP Problem and RL

The MDP consists of four fundamental components: the state space $\mathcal{S}$, the set of possible actions $\mathcal{A}$, the state transition probability $p(s'|s,a)$ and the reward probability distribution $q(r|s,a)$. Its primary characteristic is the 'future independence of the past given the present,' [24] indicating that the current state encompasses all information regarding past states.

During the operation of AUS, the efficiency of the AUS at time period $t$ depends on the current environmental observations and the system's operational actions. The energy level of the ESS $E_{t+1}$ at the next time period $t+1$ is solely determined by the current energy level $E_t$ and the charging and discharging power $c_t/d_t$ at time period $t$, independent of the previous states and actions. Consequently, the sequential decision associated with AUS energy management can be regarded as an MDP.

Recently, RL has served as a powerful machine learning algorithm to solve MDP, which seeks to maximize cumulative rewards by optimizing the agent's behavioral decisions in a stochastic environment [25]. In RL, the agent (i.e., the learner and decision maker) interacts with the environment by taking a series of actions, which yield a certain reward $r$ and cause the environment to transition to a new state $s$. The agent strives to obtain the highest possible total reward (i.e., the maximum operational efficiency of AUS) by learning how to select the optimal policy of actions that maximizes the expected cumulative reward over time.

As depicted in Figure 4, the state of AUS and real-world environmental conditions are considered as the environment in RL for solving MDP, while the AUEMS represents the agent. The AUEMS agent observes the current environmental state $s_t$ and takes actions accordingly. Subsequently, the environment transitions to the next time period, where the state changes to $s_{t+1}$ and a reward $r_t$ is returned.

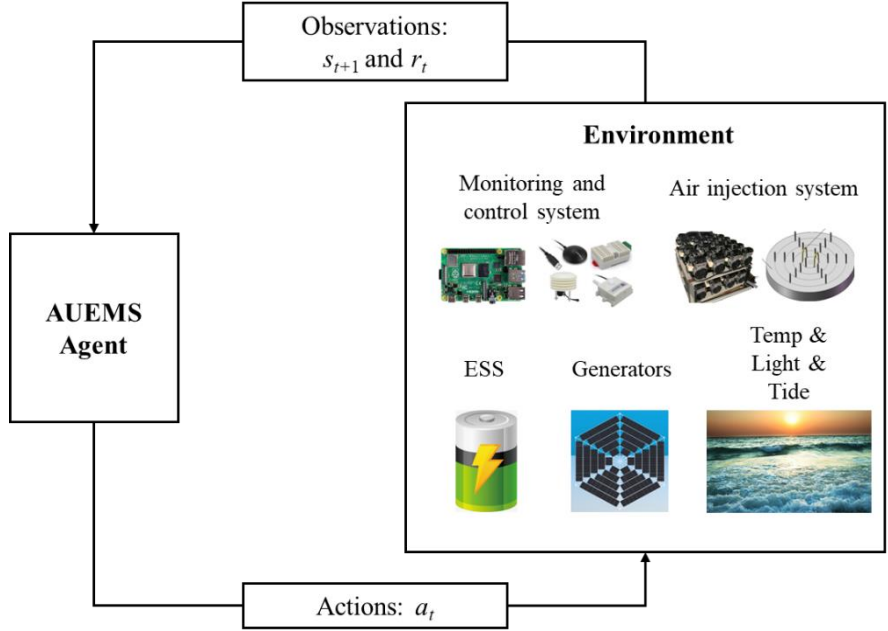

Figure 4. The agent-environment interaction in the MDP of AUS energy management.

(1) System states

The agent's decision-making process relies on the current observation $s_t$. In this study, we defined the observation consists of distinct types of information, including the energy level of the ESS $E_t$, the seawater temperature $T_t$, the light intensity $I_t$, the tidal height $Z_t$, the tidal current velocity $u_t$, and the time $t$, i.e.,

$$s_t = (E_t, T_t, I_t, Z_t, u_t, t) \tag{17}$$

Notably, the inclusion of time as an aspect of the state enables the AUEMS agent to adapt to time-dependent activities, such as time-varying tidal change and the remaining time of an aquaculture period.

(2) Action space

Regarding the AUS consisting of a set of air compressors and nozzles, it operates by injecting air into each underwater nozzle through its corresponding compressor and air pipe. Each compressor can be operated either in an "ON" or "OFF" state. Thus, the action space $\mathcal{A}$ is defined as the set of all possible combinations of compressor

states:

$$\mathcal{A} = \{\mathcal{A}_1, \mathcal{A}_2, ..., \mathcal{A}_M \mid \mathcal{A}_i \in \{0,1\}\} \tag{18}$$

where "0" or "1" indicates that the compressor is off or on, respectively. However, as the number of compressors and nozzles increases, the dimensionality of the action space surges exponentially, leading to a significantly increase in the training time and reducing the algorithm's overall performance [25]. To address this issue, we adopted a strategy that involved controlling only the number of compressors turned on $m_t$ during each time period, i.e.,

$$\mathcal{A} = \{0, 1, 2, ..., M\} \tag{19}$$

The aforementioned approach has facilitated the reduction in complexity of the MDP problem and mitigated the adverse effects caused by the 'dimensional catastrophe' phenomenon. It is also worth noting that the upwelling produced by the upstream nozzle persists on the surface of the target sea area for a longer duration than those produced by the downstream nozzle [19]. Therefore, to simplify the energy management strategy, the compressor corresponding to the upstream nozzle at time period $t$ is always switched on after determining the number of compressors to be switched on.

(3) Reward

After AUEMS agent performs action $a_t$, the system state changes from $s_t$ to $s_{t+1}$ and a reward $r_t$ is awarded [25]. Since the goal of AUEMS is to maximize system efficiency while ensuring system operation stability, we receive a reward based on the transport volume of bottom water $V_t$ at time period $t$ and penalize actions that waste energy.

$$r_t = \begin{cases} -0.1, & \text{if } a_t > 0 \text{ and } V_t = 0 \\ \beta \cdot f_{2,t} \cdot V_t(s_t, a_t), & \text{otherwise} \end{cases} \tag{20}$$

where reward scaling factor $\beta$ is introduced to limit the reward to a range (e.g., [0, 1]), which can effectively address the problems of saturation and inefficient learning in RL [26].

(4) State-action value function

When the AUEMS agent controls the ESS and air injection system at time period *t*, this agent aims to maximize the expected return it will receive in the future. Under a defined control strategy $\pi$, the return is defined as the sum of discounted rewards [25], i.e.,

$$Z_\pi = \sum_{i=0}^{\infty} \gamma^{i-1} r_{t+i} \tag{21}$$

where the discount factor $\gamma$ ($0 < \gamma < 1$) determines the importance of future rewards. The standard RL algorithm estimates the expected value of $Z_\pi$ in state *s*, known as the state-value function $V_\pi(s)$ or that of choosing a specific action $Q_\pi(s, a)$.

$$Q_\pi(s,a) = \mathbb{E}_\pi \left[ Z_\pi(s,a) \right] = \mathbb{E}_\pi \left[ \sum_{i=0}^{\infty} \gamma^{i-1} r_{t+i} \mid s_t = s, a_t = a \right] \tag{22}$$

$$V_\pi(s) = \mathbb{E}_{a \sim \pi(s)} \left[ Z_\pi(s) \right] = \mathbb{E}_{a \sim \pi(s)} \left[ Q_\pi(s,a) \right] \tag{23}$$

Moreover, the action advantage function $A_\pi(s, a)$ measures the value of selecting a specific action *a* in state *s*, which obtains a relative measure of the importance of each action by decoupling the state-value from the action-value function.

$$A_\pi(s,a) = Q_\pi(s,a) - V_\pi(s) \tag{24}$$

In the distributional RL, the value distribution (i.e., the probability distribution of $Z_\pi$) plays a central role and replaces the value function [27]. The value function $Z_\pi(s, a)$, as the expectation of the value distribution, encompasses all sources of stochasticity inherent in the value distribution [28]. Additionally, the value distribution is estimated directly to capture the stochasticity of the intrinsic returns of the MDP [29].

# 3 DRL-Based Energy Management System for Artificial Upwelling

## 3.1 Q-learning Algorithm

Solving the MDP problem detailed in Section 2.2 is equivalent to searching for the optimal policy $\pi^*$, which represents the most effective sequence of actions for the AUS to maximize system energy efficiency. Among the various RL algorithms available, the Q-learning algorithm is a commonly used model-free algorithm in decision-making processes that seek to maximize the expected cumulative reward [30].

In the Q-learning algorithm, the estimation of optimal action-value function $Q_\pi^*(s, a)$, i.e., $max_\pi Q_\pi(s, a)$, is progressively improved step by step using the Bellman optimality operator $\mathcal{T}^*$ [31], which utilizes dynamic programming approach thus significantly speeding up the learning process. Similarly, the value distribution under the optimal strategy $\pi^*$ can be calculated using the distributional Bellman optimal operator $\mathcal{T}^*$ via dynamic programming [27].

$$\begin{aligned}
\mathcal{T}^*Q(s,a) &= r(s,a) + \gamma \cdot \max(Q(s',a')), \\
\mathcal{T}^*Z(s,a) &\overset{D}{:=} r(s,a) + \gamma Z\left(s', \arg\max_{a' \in A} \mathbb{E}[Z(s',a')]\right), \\
s' &\sim P(\cdot \mid s,a), a' \sim \pi(\cdot \mid s'),
\end{aligned} \qquad (25)$$

where $Y \overset{D}{:=} U$ means that the random variables $Y$ and $U$ follow the same probability distribution.

The traditional mechanism of traditional Q-learning algorithms is to construct a Q-table [32] in which the Q-value $Q(s, a)$ of each state-action pair is updated in each iteration until the convergence condition is satisfied. However, as the state-action space expands, the complexity of the algorithm grows exponentially, also leading to the so-called "dimensional catastrophe" [24]. To mitigate this issue, recent research has leveraged deep neural networks as non-linear function approximators to estimate Q-values instead of Q-tables [16].

(1) *Deep Q Network* (*DQN*)

In 2015, V. Minh et al. proposed the DQN algorithm [16], which utilizes deep neural networks to parameterize the approximation function $Q(s, a; w_i)$ of the optimal action-value function $Q_\pi^*(s, a)$, where $w_i$ represents the Q-network parameter at the $i$-th iteration. To ensure the stability of DRL, experience replay [33] and target Q-network [16] are the most commonly employed techniques.

Experience replay involves storing the agent's experience $e_t = (s_t, a_t, r_t, s_{t+1})$ at each iteration in the dataset $D_t = \{e_1, …, e_t\}$. During the learning process, experience tuples (or mini-batches) are randomly sampled from the dataset and applied to Q-learning updates. The Q-learning update for *i*-th iteration is typically computed using the following loss function [16]:

$$L_i(\theta_i) = \mathbb{E}_{(s,a,r,s') \sim U(D)} \left[ \left( r + \gamma \max_{a'} Q(s', a'; w_i^-) - Q(s, a; w_i) \right)^2 \right] \quad (26)$$

$$y^{DQN} = r + \gamma \max_{a'} Q(s', a'; w_i^-) \quad (27)$$

where $w_i^-$ represents target Q-network parameters used in the *i*-th iteration and $w_i^-$ are updated to Q-network parameters $w_i$ only every $C$ step, and remain constant before each update.

(2) *Dueling Deep Q Network* (*Dueling DQN*)

The enhance the convergence speed of DQN, Wang et al [34] presented a novel neural network architecture, known as Deep Dueling Network, which improves policy evaluation in the presence of numerous actions of comparable value.

In the deep dueling network, the action-value function is decoupled by a sequence of two fully connected layers (or streams), estimating the state-value $V_\pi(s)$ and the action advantage $A_\pi(s, a)$, respectively. These values are then combined at the output layer. To obtain the unique state-value *V* and advantage *A*, an approximation of the advantage function is assumed to have zero advantage in choosing an action. Therefore, the output of the Dueling DQN is the combination of the two streams:

$$Q(s, a; w, \alpha, \beta) = V(s; w, \beta) + \left( A(s, a; w, \alpha) - \frac{1}{|\mathcal{A}|} \sum_{a'} A(s, a'; w, \alpha) \right) \quad (28)$$

where *α* is the parameters of the state value stream and *β* is the parameters of the action advantage stream.

(3) *Deep Q-Network with Quantile Regression* (*QR-DQN*)

To further improve the robustness of the Q-learning algorithm to hyperparameter changes and environmental noise, Bellemare et al [27] proposed a distributional Q-learning algorithm called C51. Unlike traditional Q-learning, which estimates only the mean of the return, C51 models the entire distribution of the return. Specifically, it learns the value distribution $Z_\pi(s, a)$ instead of the optimal action value function $Q_\pi^*(s, a)$. The objective of the distributional Q-learning algorithm is to minimize a statistical distance between the learned distribution and the true distribution of the return.

$$\sup \text{dist}(\mathcal{T}^* Z(s,a), Z(s,a)) \tag{29}$$

where dist(X, Y) denotes the distance between random variables X and Y, which can be measured using metrics such as KL scatter [27], p-Wasserstein metric [35], etc. The p-Wasserstein metric between two real-valued random variables U and V is given by the following equation

$$W_p(U,V) = \left( \int_0^1 \left| F_V^{-1}(\omega) - F_U^{-1}(\omega) \right|^p d\omega \right)^{1/p} \tag{30}$$

where $F_U$ and $F_V$ are the cumulative distribution functions of U and V, respectively. Despite the fact that the distributional Bellman operator $\mathcal{T}^*$ is a strict contraction of the p-Wasserstein distance [27], the C51 algorithm did not itself minimize the Wasserstein metric. To address this limitation, Dabney et al [35] proposed the QR-DQN algorithm based on quantile regression theory [36], which uses uniformly distributed measures to estimate quantiles of the target distribution (as shown in Figure 5). The quantile Q-values of each action can be calculated as follows:

$$Q(s,a) = \sum_{i=1}^{N} q_i \theta_i(s,a) \tag{31}$$

where N is the number of quantile bits, $\theta$ is the support of the distribution. The aim of QR-DQN is to minimize the 1-Wasserstein distance (i.e., $p = 1$ for the p-Wasserstein distance). To train the neural network in QR-DQN, the quantile huber loss of quantile regression is used:

$$\rho_\tau^\kappa(u) = \begin{cases} \frac{1}{2} u^2 \left| \tau - \delta_{\{u<0\}} \right|, & \text{if } |u| \leq \kappa \\ \kappa \left( |u| - \frac{1}{2}\kappa \left| \tau - \delta_{\{u<0\}} \right| \right), & \text{otherwise} \end{cases} \tag{32}$$

where $\tau$ is the center of the equidistributed *y*-axis value of the cumulative distribution function, which can be expressed by the following equation:

$$\tau_i = \frac{2(i-1)+1}{2N}, i=1,\ldots,N \tag{33}$$

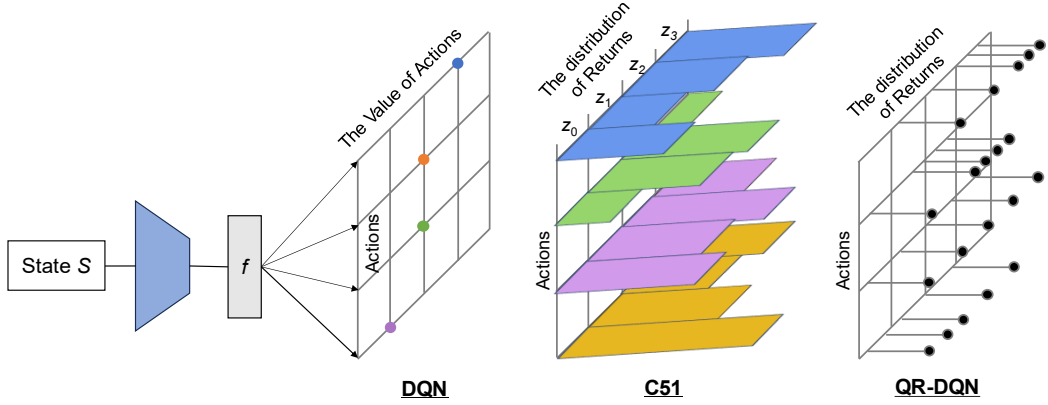

Figure 5. The difference between DQN, C51, and QR-DQN.

## 3.2 DRL for Decision-making

To solve the efficiency optimization and MDP problem of the AUS defined in Section 2.2, we present a novel DRL-based energy management framework by combining QR-DQN with deep dueling network, as illustrated in Figure 6. Algorithm 1 elaborates the detailed process of the proposed energy management algorithm, which comprises an outer loop and an inner loop. The outer loop governs the number of training episodes, while the inner loop executes AUS control during each time period within a single training episode. The termination of each episode occurs either when the time period of each episode reaches the maximum period $T$ or the energy level of ESS $E_t$ at time period $t$ is less than $E_{min}$.

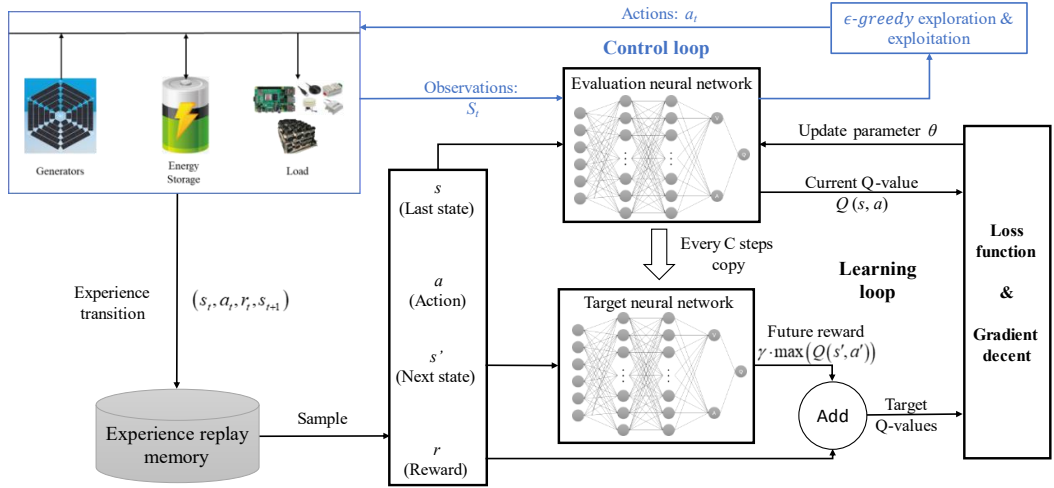

Figure 6. Artificial upwelling energy management framework based on DRL.

At each time step, AUEMS agent adopts the $\epsilon\text{-}greedy$ strategy ($\epsilon \in [0, 1]$) to select actions, aiming to balance the exploration and exploitation process in the state space and achieve optimal results. This approach enables the agent to randomly select actions from the action space $\mathcal{A}$ with a probability of $\epsilon$, while also taking actions according to the Q-network $Q(s, a; \theta)$ with a probability of $1-\epsilon$, which means that the agent chooses the action related to the largest current Q-value:

$$a_t = \arg\max_{a \in \mathcal{A}} \mathbb{E}\left[Q(s_t, a; w)\right] \tag{34}$$

By balancing exploration and exploitation, the agent can explore the action-state space with some degree of randomness, while also avoiding being completely random. After selecting an action, the agent receives a reward $r_t$, observes the next state of the environment $s_{t+1}$, and stores the experience tuple $(s_t, a_t, r_t, s_{t+1})$ into the experience replay buffer. In the subsequent training process, random samples from the experience replay buffer will be used to update the Q-network parameters $w$. This process is repeated until the end.

**Algorithm** 1 DRL-Based Energy Management Algorithm for AUS

Initialize replay memory $D$ to capacity $N$

Initialize the action-value function $Q$ with random weights $w$

Initialize the target action-value function $\hat{Q}$ with random weights $w^- = w$

**For** episode 1, *M* **do**

  Initialize sequence

 **For** $t =1$, *T* **do**

    Choose the action $a_t$ for the current state $s_t$ by using $\varepsilon\text{-}greedy$ policy

    Perform action $a_t$ and observe reward $r_t$ and next state $s_{t+1}$

    Store transition $(s_t, a_t, r_t, s_{t+1})$ in *D*

    Sample random minibatch of transitions $(s_j, a_j, r_j, s_{j+1})$ from *D*

$$\text{set } T\theta_j = \begin{cases} r_j \\ r_j + \gamma\theta_j\left(s', \arg\max_{a'\in A}\sum_j q_j\theta_j(s',a')\right) \end{cases}$$

    if episode terminates at step *j*+1 otherwise Perform a gradient descent step on

$\sum_{i=1}^{N} \mathbb{E}_j\left[\rho_{\tau_i}^{\kappa}\left(T\theta_j - \theta_i(s_j, a_j)\right)\right]$ with respect to the network parameters *w*

    Every *C* steps reset $\hat{Q} = Q$

 **End**

**End**

## 3.3 Neural Network Design

The neural network architecture developed to estimate the quantile Q-values adopt the widely employed settings [26,37]. As shown in Figure 7, the network architecture comprises an input layer and two hidden layers, in which the input layer receives in the state vector *s*, including significant features pertaining to the marine environment and AUS. Since the value range of each feature may vary greatly, we utilized the Z-score normalization method [38] to pro-process the state vector *s*, and scaled their values to make our proposed algorithm achieve a more stable learning process.

$$x' = \frac{x - \text{mean}(x)}{\text{std}(x)} \tag{35}$$

where *x* represents the features in the state *s*, and the mean and standard deviation values of each feature were estimated from historical observations. The two hidden layers that follow the input layer employ Rectified Linear Unit (ReLU) as the activation function, with each layer consisting of 64 × N neurons. A fully connected layer *V*(*s*; *w*, *β*) is the employed to approximate the state-value function, and another fully connected layer *A*(*s*; *w*, *α*) is used to approximate the action advantage function. Finally, the quantile Q-value is obtained by combining the outputs of the two fully connected layers via Eq. (30).

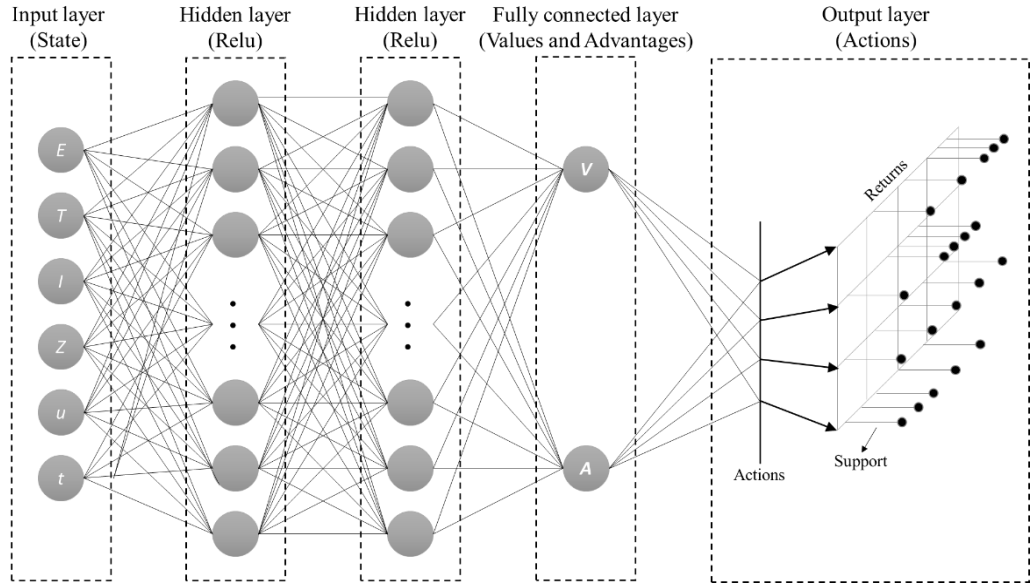

Figure 7. Structure of Neural network utilized in artificial upwelling energy management. (The example shows a neural network with 4 actions, and the actual number of actions is changed with the number of air compressors of AUS.)

## 4 Results

### 4.1 Simulation Setup

To evaluate the performance of our DRL-based energy management algorithm presented in Section 3, we used weather data obtained from NASA's Power Data Archive (https://power.larc.nasa.gov/) and tide data from the China National Oceanic Data Center (http://mds.nmdis.org.cn/, as shown in Figure 8) to simulate the test environment in Aoshan Bay. Since the primary growth period of kelp in China is typically between February to June [39], we selected data ranging from February 1st to June 1st for the years 2016 to 2018, which served as the training set for the neural network model. Subsequently, we used data from February 1st to June 1st of 2019 to evaluate the algorithm's performance.

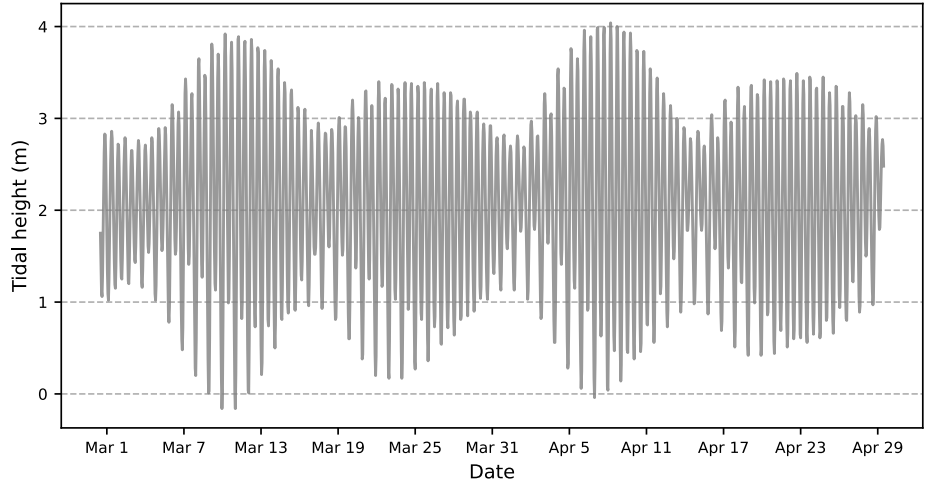

Figure 8. Variation of tidal height from February 1st to June 1st in 2019. (Tide data are obtained from http://mds.nmdis.org.cn/)

Furthermore, we considered several parameters related to the studied AUS in Aoshan Bay [10]. The photovoltaic system's installed capacity $P_{AZ}$ = 48 kW, with a total output power of 36 kW. The maximum storage capacity of ESS $E_{max}$ = 86.4 kWh, and the charge and discharge efficiency $\eta_c = \eta_d$ = 0.95. The essential load power of AUS was assumed to be constantly 0.2 kW. The air injection system comprised 16 air compressors, each have a rated power $P_0$ = 1 kW and an air injection rate $Q_0$ = 6 m³/h. As shown in Figure, there are 16 nozzles arranged at equal intervals of 4m in the center of a 120×70 m² aquaculture area. The related environmental and seaweed growth parameters in Aoshan Bay demonstrate as follows: the density difference between the upper and bottom water layers $\Delta\rho$ = 0.6 kg/m³ [13], the thickness of the surface layer $h$ = 2 m, the optimal light intensity for the growth of seaweed growth $I_{opt}$ = 180 W/m² [39], the optimum growth temperature $T_{opt}$ = 10 °C [39].

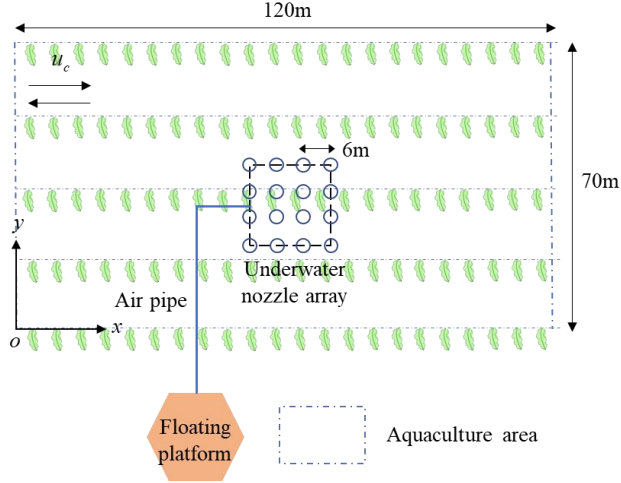

Figure 9. Schematic diagram of the experimental area and the layout of the AUS in Aoshan Bay.

For the simulation, we adopted parameter settings akin to those detailed in the literature for training the AUEMS agent [35]. Specifically, the learning rate $\alpha = 0.00025$, the exploration rate $\epsilon$ was linearly annealed from 1 to 0.01 in the first 10% iterations and remained constant in subsequent iterations, and the number of quantile supported $N = 200$. The main parameter settings are shown in Table 1.

Table 1. Simulation parameters setup

| | | | |
|---|---|---|---|
| $H_{test}$ | 2,880 hours | $\Delta t$ | 1 hour |
| $E_{max}$ | 86.4 kWh | $E_{min}$ | 4 kWh |
| $P_{AZ}$ | 48 kW | $p_0$ | 1 kW |
| $\gamma$ | 0.99 | $\epsilon_{min}$ | 0.01 |
| $C$ | 10,000 | $N$ | 200 |
| batch size | 32 | buffer size | 1,000,000 |
| Optimizer | Adam | $\alpha$ | 0.00025 |

## 4.2 Simulation results

In this section, we presented the results of our experiments, evaluating the performance of our proposed DRL algorithm (QR-DQN with the deep dueling network) in managing the energy schedule of AUS over a duration of 2,880 hours. Our methodology was compared with other DRL algorithms, including QR-DQN devoid of a deep dueling network, as well as those without Distributional DQN and DDQN (Double DQN [40] with the deep dueling network) and a rule-based greedy strategy [13]. It's noted that we used photosynthetic factor $f_{2,t}$ to represent the consideration of kelp growth in [13]. When $f_{2,t}$ is less than the minimum photosynthetic factor $f_{min}$, the air injection system does not work. Additionally, we examined the performance of our algorithm in diverse tidal conditions, including a simplified tide model in Aoshan Bay [11] and realistic tide.

Figure 10 illustrates the learning process and convergence of the proposed algorithm compared to other DRL algorithms under the simplified tide model. Based on the sampled data, this model simplifies the regular semi-diurnal tide in Aoshan Bay as a sinusoidal function that represents the variation of sea level ranging from 7.1 m to 9.5 m and tidal speed ranging from 0 to 0.25 m/s over time. To generate the simulation results, we executed five independent runs of 1,000,000 steps and reported the average and standard deviation of rewards in Figure 10(a). The results demonstrate that the DRL agent could self-learn more effective strategies as it interacted with the environment, leading to a more stable running of AUS and a generally increasing average reward. Furthermore, it can be observed that the convergence speed and robustness of the proposed DRL algorithm outperformed those of other DRL algorithms.

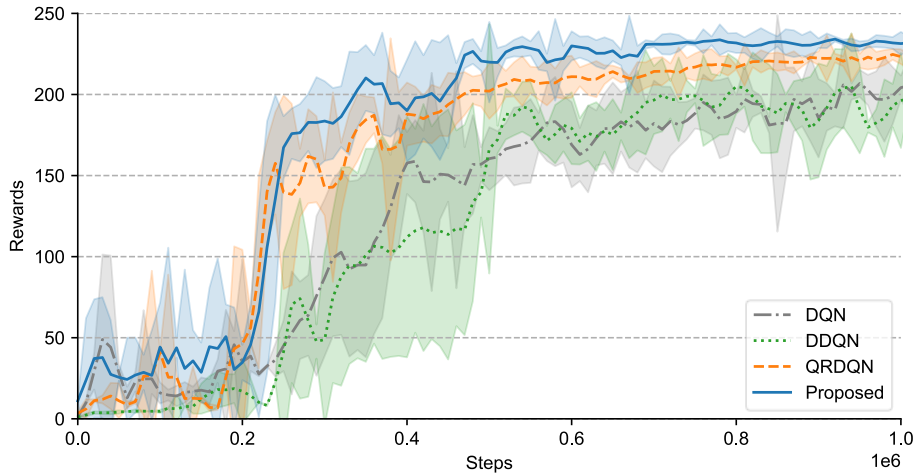

(a) rewards per 10,000 training steps.

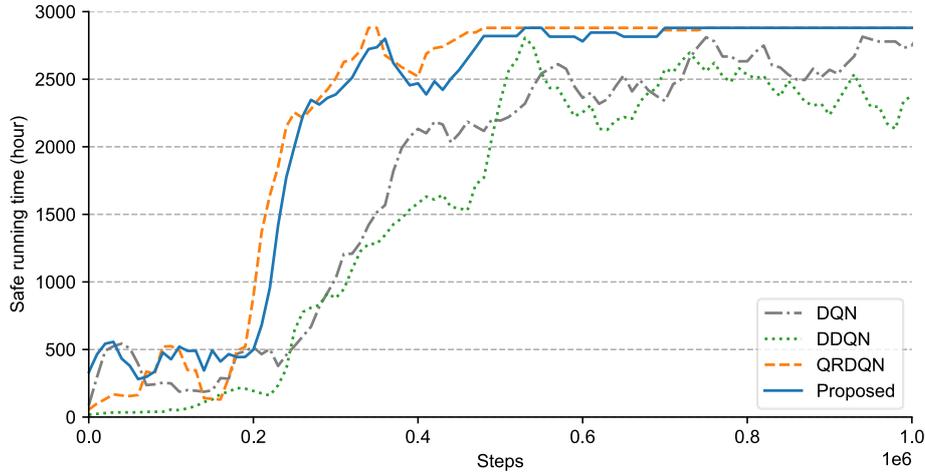

(a) safe running time per 10,000 training steps.

Figure 10. Training process and performance comparisons among four DRL algorithm.

Specifically, after sampling and training 350,000 steps on average, the distributional DRL algorithm (i.e., QRDQN and our proposed algorithm) can firstly achieve securing the running of AUS during the 2880-hour aquaculture period. This is because the proposed algorithm could first learn an efficient strategy to balance the power supply for essential loads and air injection system, ensuring the stable operation of the system (as shown in Figure 10b). After 1,000,000 steps of sampling, our proposed DRL algorithm could trend to converge and acquire the optimal strategy that maximized the efficiency of AUS, achieving a maximum reward better than other DRL algorithms. Furthermore, the standard deviation of results generated by the proposed algorithm was smaller than other algorithms, demonstrating its superior robustness. These findings reveal that by estimating the value distribution directly, the distributional DRL algorithms can capture the full range of possible outcomes and make informed decisions to maximize the expected return, improving the system's performance and efficiency.

Table 2 provides a comprehensive performance comparison of various DRL algorithms and the greedy algorithm with different $f_{min}$ in terms of total air injection volume, total transport bottom water volume, average value of photosynthetic factor, total energy wastage, and total efficiency. The obtained results reveal that our proposed algorithm outperformed the other algorithms by a significant margin, achieving an optimal performance of 247,315, which represents an impressive 42.34% and 11.76% improvement over the greedy algorithm ($f_{min} = 0$ and $f_{min} = 0.2$), respectively. The superiority of the proposed DRL algorithm is evident from its remarkable ability to increase the average photosynthetic limit to 0.393 compared to other DRL algorithms while maintain a relatively low energy consumption and energy wastage.

Table 2. Performance comparison of DRL algorithms and Baseline algorithm

|  | Total air injection volume (m³) | Total transport volume (m³) | Average photosynthetic factor | Total Energy wastage (kW) | Total Efficiency |
|---|---|---|---|---|---|
| Greedy ($f_{min} = 0$) | 35,286 | 746,325 | 0.266 | 9 | 173,759 |
| Greedy ($f_{min} = 0.2$) | 22,152 | 423,191 | 0.538 | 2150 | 221,562 |
| DQN | 27,720 | 562,832 | 0.373 | 1241 | 234,955 |
| DDQN | 34,032 | 653,159 | 0.341 | 212 | 237,972 |
| QRDQN | 32,190 | 670,116 | 0.323 | 520 | 244,355 |
| Proposed | 32,388 | 672,084 | 0.393 | 452 | 247,315 |

(Note: the average value of photosynthetic factor is calculated by $\sum_{t=1}^{T}\{f_{2,t}/T \mid a_t > 0\}$)

As depicted in Figure 11, during the initial period of aquaculture cycle (i.e., 1st February to 15th March), the greedy algorithm ($f_{min} = 0.2$) demonstrated a comparable energy management strategy and reward with our proposed algorithm. However, as environmental conditions, such as light intensity and temperature, evolve over time, the limitations of the greedy algorithm become evident. This algorithm relies solely on the available energy and photosynthetic limits of the system, rendering it inadequate in dynamically adapting to changing environmental states. Consequently, energy wastage and diminished system efficiency become apparent. In contrast, our proposed DRL algorithm surpasses the performance of the greedy algorithm owing to its ability to dynamically manage energy allocation based on its experience self-learned from historical data. The observed improvement in the performance of our algorithm can be attributed to its capacity to achieve an optimal balance between conserving energy for potential high-reward work instances and unnecessary energy consumption. This flexibility and efficiency in energy scheduling culminate in a notable reduction in energy consumption while concurrently bolstering system performance.

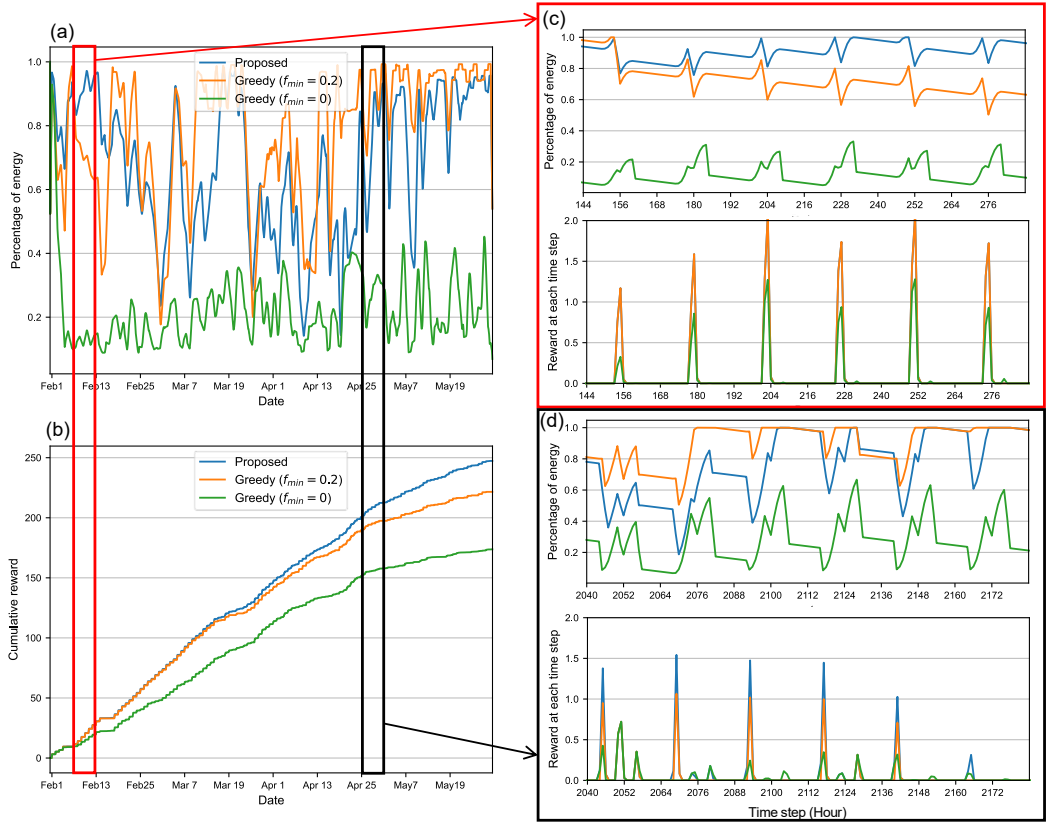

Figure 11. Simulation results associated with the ESS and rewards. (a) the percentage of energy stored in the ESS. (The result was smoothed using the move average method [41].) (b) the reward. (c) energy and reward from 6th Feb to 12th Feb. (c) energy and reward from 1th May to 6th May.

Figure 12 provides additional insight into the action selection process of the proposed DRL algorithm. The state-action distribution results reveal the Q-values of each action under different states. Specifically, the Q-values are computed based on the expected value of each state-action distribution, and subsequently, actions are chosen based on their corresponding Q-values. The selected actions are highlighted in red in the graph.

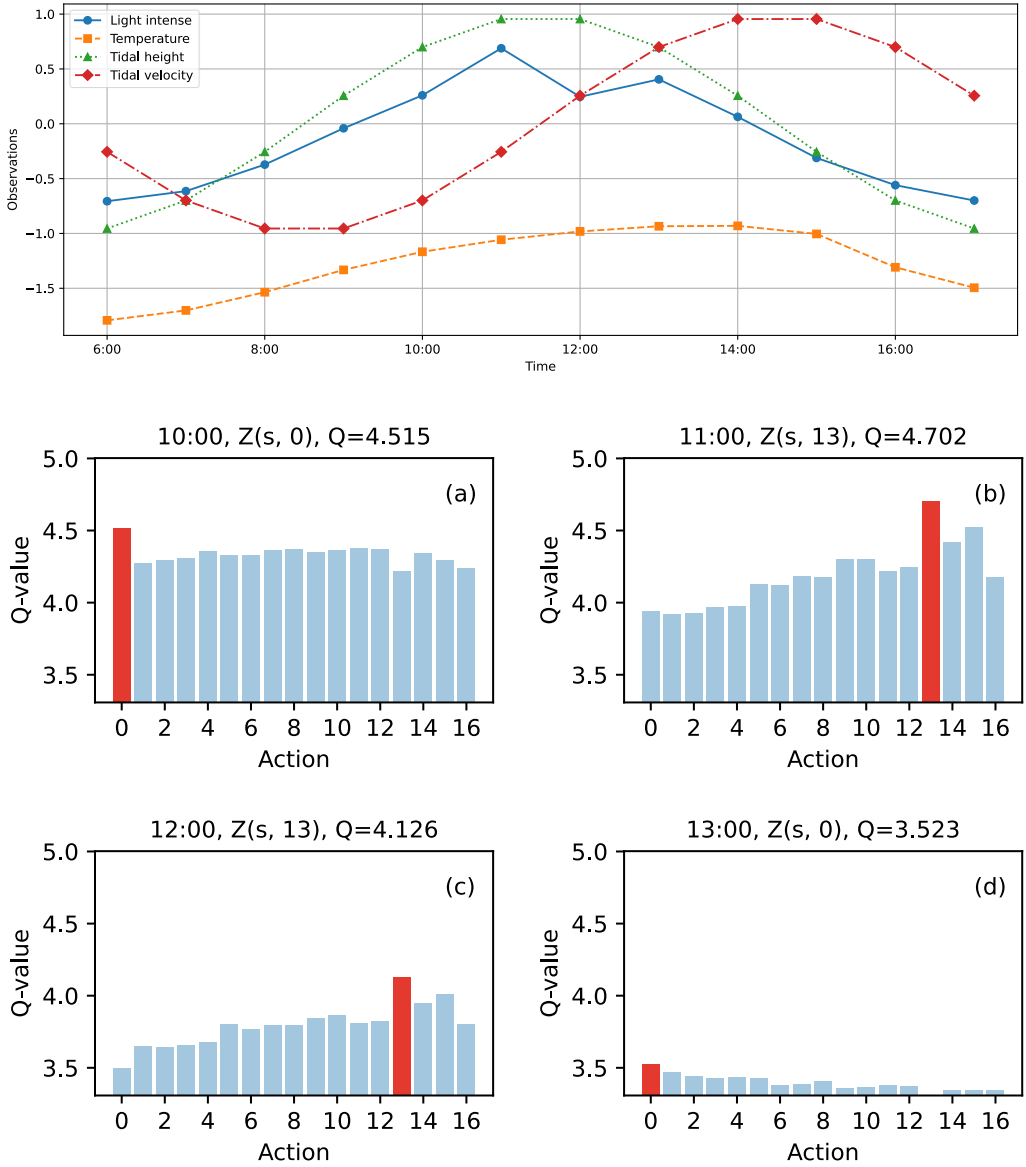

Figure 12. An example sequence of state-action distribution Z(s, a).

To further validate the superiority of our proposed DRL algorithm, we conducted an empirical evaluation was to assess the performance of various algorithms in the simulated environment that incorporated realistic tidal data (see Figure). Since the level of complexity for state space increased in this situation, we allocated enhanced computational resources during the learning process. Specifically, we conducted 1.5 million samples per learning iteration, with mini-batches consisting of 64 samples each. To gauge the effectiveness of each DRL algorithm, the performance of the greedy algorithm ($f_{\min} = 0$) was employed as a baseline. The improvements achieved by the different DRL algorithm compared to the baseline are depicted in Figure 12. The results show that our proposed DRL algorithm attained optimal performance improvement by 20% within the designated 1.5 million timesteps. In contrast, the impact of regulating the minimum photosynthetic factor $f_{\min}$ in the greedy algorithm is comparatively insignificant, resulting in a modest enhancement of merely 5%.

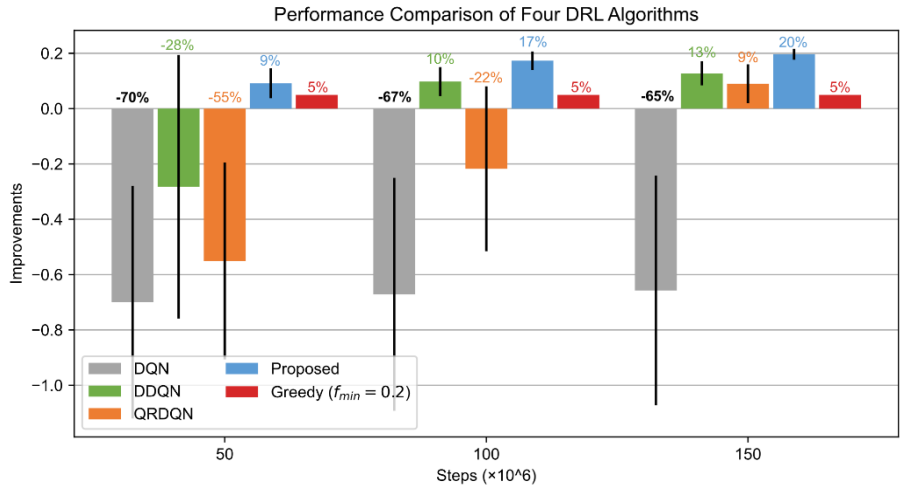

Figure 13. Performance comparison of four DRL algorithms and the greedy algorithm in the simulated real ocean environment

Furthermore, the DQN algorithm encounters significant challenges in learning efficient policies and achieving convergence due to the vast number of state-action pairs, consequently leading to inferior performance compared to other algorithms. While the DDQN and QR-DQN algorithms demonstrate respective performance improvements of 13% and 9%, they exhibited slower convergence rates and lower robustness compared to the proposed algorithm, as indicated by larger standard deviations. These findings underscore the effectiveness of our proposed DRL algorithm, which effectively addresses the issue of overestimation caused by the optimizer, specifically stochastic gradient descent. This is achieved through the integration of distributional DRL with deep dueling neural networks, which decouple the neural network into two distinct streams. This approach mitigates the impact of overestimation and contributes to the superior performance of our proposed algorithm.

These findings highlight the potential for developing more dependable and efficient energy management strategies for AUS. By leveraging advanced algorithms like DRL, the seaweed cultivation and efforts towards carbon sequestration can experience significant enhancement.

## 5 Conclusion

This study proposes a novel approach to energy management of AUS by combining QR-DQN, a distributional RL algorithm, with deep dueling neural networks. This approach considers the effects of tide, light, and temperature on AUS efficiency as key factors in controlling the air injection without prior experience and hydrodynamic and biological growth models. Simulation results based on realistic data demonstrate the effectiveness and robustness of the proposed algorithm, which outperforms non-distributional methods that deterministically estimate the expected future rewards. By using the deep dueling network to decouple the value of the action and the value of the state, a faster convergence speed and better performance of the proposed algorithm than the QR-DQN are obtained.

Overall, this study confirms the potential of DRL algorithms in planning AUS operations and suggests new research directions for energy management and RL. In future work, we plan to investigate the scalability and robustness of the proposed algorithm for large-scale AUS. We also aim to explore the possibility of incorporating other renewable energy sources, such as wind power and wave power, into AUS and our energy management strategy. These endeavors will provide further insights into the potential of distributional reinforcement learning algorithms for optimizing AUS performance and energy efficiency.